\documentclass[letterpaper, 10 pt, conference]{Config/ieeeconf}

\IEEEoverridecommandlockouts
\usepackage[utf8]{inputenc}
\usepackage{amsmath}

\usepackage{amsfonts}
\usepackage{amsthm}
\usepackage{amssymb}
\usepackage{graphicx}
\usepackage{booktabs}
\usepackage{hyperref}
\hypersetup{hidelinks}
\usepackage{url}
\usepackage{comment}
\usepackage{algorithm}
\usepackage{algpseudocode}
\usepackage{cite}
\usepackage{dblfloatfix} %warning ma mette la figura in basso

\usepackage{xcolor}

\title{Compliant Blind Handover Control for Human-Robot Collaboration} %

\author{Davide Ferrari, Andrea Pupa and Cristian Secchi% <-this % stops a space
\thanks{D. Ferrari and C. Secchi are with the Department of Sciences and Methods of Engineering, University of Modena and Reggio Emilia, Italy {\tt\small\{davide.ferrari95, andrea.pupa, cristian.secchi\}@unimore.it}}} 

\newcommand{\q}{\mathbf{q}}
\newcommand{\dq}{\dot{\mathbf{q}}}
\newcommand{\ddq}{\ddot{\mathbf{q}}}
\newcommand{\x}{\mathbf{x}}
\newcommand{\dx}{\dot{\mathbf{x}}}
\newcommand{\ddx}{\ddot{\mathbf{x}}}
\newcommand{\uin}{\mathbf{u}}
\newcommand{\R}{\in\mathbb{R}}
\newcommand{\traj}{\q_{des}(t)}
\newcommand{\trajs}{\q_{des}(s(t))}
\newcommand{\dtraj}{\dot{\q}_{des}(t)}
\newcommand{\dtrajs}{\q'_{des}(s(t))}

\newcommand{\n}{\mathbf{n}}

\begin{document}
    
    \maketitle
    
    \begin{abstract}

        This paper presents a Human-Robot Blind Handover architecture within the context of Human-Robot Collaboration (HRC). The focus lies on a blind handover scenario where the operator is intentionally faced away, focused in a task, and requires an object from the robot. In this context, it is imperative for the robot to autonomously manage the entire handover process. Key considerations include ensuring safety while handing the object to the operator's hand, and detect the proper timing to release the object. The article explores strategies to navigate these challenges, emphasizing the need for a robot to operate safely and independently in facilitating blind handovers, thereby contributing to the advancement of HRC protocols and fostering a natural and efficient collaboration between humans and robots.

    \end{abstract}
    
    % Input Chapters
    
\section{Introduction}\label{sec:intro}

    In collaborative robotics, the handover represents a fundamental aspect, embodying the most classic form of interaction among team members - the transfer of an object. Formally, a handover is a \textit{"joint action between a giver and a receiver"} \cite{sebanz06TCS}. It is a type of social interaction where \textit{"two or more individuals coordinate their actions in space and time to bring about a change in the environment"}. While both agents share the goal of transferring the object, their specific objectives differ throughout the interaction \cite{Mason05EBR}. The giver aims to present the object appropriately, hold it stably during the physical handover, and release it safely to the receiver. On the other hand, the receiver's objectives include acquiring the object, stabilizing the grasp, and subsequently using the object for its intended task. This interaction spans various Human-Robot Collaboration (HRC) contexts, such as industrial \cite{Zeng22IJRR, Ortenzi22ACM, Male22SMC}, medical \cite{Choi09ROMAN, Nowak22ARSO}, domestic \cite{Detry17IROS, Lastrico23IROS}, etc. 

    It can be divided in two phases \cite{Ortenzi21TRO}: \textit{Pre-Handover} and \textit{Physical Handover}. The pre-handover phase begins with a request for an object, communicated from one agent to another. Communication plays a crucial role in this phase and can occur in various ways to coordinate the what, when, and where of an handover. Common methods include gaze, pose, oral cues, but also motions or gestures where the giver clearly indicates the intent to hand over the object. Similarly, the grasp and offering of the object also provide cues regarding the intent to hand over. Literature has extensively explored various aspects of the pre-handover phase, including synchronization and coordination between the actors \cite{Strabala13JISC}, the effects of vocal communication in joint action \cite{Masumoto14EBR}, and communication involving shared attention, gaze, visual and haptic cues \cite{Moon14ACM, Costanzo23ROMAN, Newbury22ROMAN}. Other works have focused on grasp planning and how to place the object in the receiver's hand \cite{Chan15IROS, Bohg14TRO, Chan20IRSR, Ortenzi20Frontiers}, attempting to replicate the strategies used during a human-human handover or on the perception of the object, the hand (self and partner), and the partner’s full-body motion \cite{Hasson19CVPR, Yang21ICRA, Kokic17Humanoids, Rosenberger21RAL}.
    On the other hand, the physical handover phase involves the physical interaction between the giver and the receiver, along with the transfer of the object. Both players are physically and cognitively engaged in this phase, which begins when the receiver's hand makes contact with the object. During this phase, control over the object's possession and stability is transferred from the giver to the receiver. The giver may use vision and force feedback to gauge the receiver's grasp on the object, determining when to release it for a smooth transition. Timing coordination is crucial, as an early release can cause the object to fall, while a late release can result in higher interaction forces \cite{Chan12HRI}. Literature has explored strategies for an effective transition, including the use of tactile sensors, wearable gloves, or other devices to detect changes in grip force and identify the correct timing for the transition \cite{Zou17Sensors, Leal19ISMCR, Wang19THMS, Rayamane23Springer}. Other studies \cite{Chan13IJRR, Chan14ICRA, Controzzi18EBR} have focused on force-torque sensor feedback to analyze load curve transitions during object handovers.
    
    This paper specifically concentrates on the physical handover, incorporating the pre-handover phase with a voice communication architecture based on dialogues \cite{Ferrari23IRIM}, recognizing language as a form of joint action \cite{Clark96Cambridge, foster06ACM}. Since the implementation of invasive and delicate wearable or tactile sensors is often required in a controlled environment, and is more suitable for lightweight objects, this article relies solely on force feedback to recognize object transitions. Additionally, the decision was made not to employ cameras for implementing object perception through vision algorithms, to create an efficient platform with minimal necessary hardware. However, these elements may be integrated in the future to extend the proposed architecture.
    
    Notably, human involvement in the referenced papers is active during the handover, with the human approaching the robot and coordinating movements.
    % In contrast, our novel concept introduces Blind Handover, simulating a scenario where the user is engaged in a secondary task, deliberately positioned facing away from the robot, unable to divert attention, and extends a hand to the robot requesting a tool.
    In contrast, our novel concept introduces Blind Handover, simulating a situation where the user is focused on another task, deliberately facing away from the robot. As the user is unable to divert attention, he extends a hand and request a tool without looking at the robot. 
    % The robot must execute the handover procedure by tracking the user's hand and releasing the tool only when the user has full control. Given that the operator is facing away, 
    The robot must handle errors independently, ensuring safety and a compliant movement that aligns with the operator's expectations, resembling a human-human handover. This introduces the need to strictly adhere to ISO/TS 15066 safety standards as the operator lacks of visual feedback, requiring to place complete trust in the robot's execution of the task.
    Since no visual feedback is used, ensuring a robust transfer guided solely by the object load-transfer curve estimation based on force sensor readings is crucial. Furthermore, it is very likely that the operator, blindly reaching for the object, may inadvertently cause shocks or contacts while trying to perform the grasping. These disturbances must be accounted to create a robust object release strategy.
    
    % This unique challenge emphasizes the need for careful consideration of safety standards, specifically ISO/TS 15066, given the lack of visual feedback, requiring complete trust in the robot's task execution.
    
    % This setup introduces a unique challenge, necessitating careful consideration of safety standards ISO/TS 15066 as the operator lacks of visual feedback, requiring to place complete trust in the robot's execution of the task.
    
    To address these challenges, this paper proposes:
    \begin{itemize}
        % \item  An innovative safe and compliant blind handover control architecture enabling the execution of blind handover tasks while adhering to safety ISO regulations.
        \item An innovative safe blind handover control architecture that allows for the natural and user-compliant execution of blind handover tasks while adhering to ISO safety regulations.
        \item  A neural network estimating the load curve variation during the physical handover, coupled with a compliant trajectory control, facilitating a natural and robust object transfer resilient to external disturbances.
        \item  Experimental validation by comparing our proposed architecture with the state-of-the-art, using a non-compliant PD controller and a force threshold to detect the gripper's opening moment.
    \end{itemize}

    This paper is organized as follows: Section II present the Problem Statement. Section III introduces the Proposed Compliant Blind Handover Architecture, and Section IV provides insights into the Safety Layer. Section V delves into the implementation of the Force-Load Transfer Neural Network. Section VI covers Experimental Validation, Implementation Details, and Analysis of the Results. Finally, Section VII concludes the paper.

    \section{Problem Statement}\label{sec:problem}

    % \begin{itemize}
    %     \item consider a hrc blind handover scenario
    %     \item robot must pass a requested object to a operator facing away.
    %     \item NI (attualmente object pose hardcodata e human hand rilevata come punto) robot must estimate the best grasping pose and the best handover pose by tracking the pose of the human hand.
    %     \item safety ISO/TS 15066 must be ensured, compliance behavior during the handover phase must be ensured (SSM + PFL)
    %     \item robot must recognize the moment to open the gripper and release the object, ensuring the operator has the full control of the object.
    % \end{itemize}
    Consider a HRC scenario where a velocity controlled robot manipulator has to perform a blind handover task to the human operator. Thus, the $n$-DoFs collaborative manipulator can be modeled as:
    \begin{equation}
        \dq = \uin,
    \end{equation}
    where $\dq\R^n$ and $\uin\R^n$ represents the joint velocities and the controller input, respectively.
    In such a scenario, the robot has to provide the human operator with an object necessary for performing a task, e.g., a screwdriver, while the human operator is not looking at the robot itself. Thus, it is assumed that the collaborative cell is endowed with a monitoring unit that is able to track the human operator hand in real-time inside the scene. This monitoring is exploited for planning trajectories $\traj\R^n$ that moves the robot from the initial configuration $\q_{des}(t_i)=\q_i\R^n$ to a desired final configuration $\q_{des}(t_f)=\q_f\R^n$, in order to pass the object to the human operator.
    % During the collaboration, the robot behavior has to be compliant with the safety standards imposed by the ISO/TS 15066 \cite{standards}.
    Crucially, during collaboration, these trajectories must comply with safety standards outlined in ISO/TS 15066 \cite{isots}, represented by two paradigms: Speed and Separation Monitoring (SSM) and Power and Force Limiting (PFL).
    In particular, the SSM paradigm places restrictions on the maximum speed towards the operator, shaping the robot's behavior based on the distance between the robot and the human \cite{Pupa24RAL}:
    \begin{equation}
        \begin{split}
            v_{rh,SSM}(t)= &\sqrt{v_h(t)^2+(a_{max}T_r)^2+2a_{max}K(t)}\\
            &+a_{max}T_r -v_h(t),
            \label{eq:ssm}
        \end{split}
    \end{equation}
    where $K(t) = C + Z_d + Z_r-S_p(t)$, $v_{{rh}}(t)$ and $v_{h}(t)$ represent the scalar velocity of the robot towards the human operator and the scalar velocity of the human operator towards the robot, respectively. $a_{max}$ is the maximum deceleration, $T_r$ is the robot reaction time, $C$ is the intrusion distance, $Z_d$ and $Z_r$ are the position uncertainties of the human operator and the robot system, respectively, and $S_p$ represents the protective separation distance.
    The PFL paradigm, instead, imposes a maximum energy limit that can be exchanged in case of contact between the robot and a part of the operator's body.
    % , based on pain tolerance tests for each body area
    Following the guidelines of Appendix A of the ISO/TS 15066 standard \cite{isots}, it is possible to derive a maximum permissible velocity limit:
    \begin{equation}
         v_{rh,PFL}(t) = \sqrt{\frac{2E_{\max}}{\mu}} = \frac{F_{\max}}{\sqrt{\mu \cdot k}} = \frac{p_{max} \cdot A}{\sqrt{\mu \cdot k}}.
         \label{eq:pfl}
    \end{equation}
    Here, $E_{\max}$ is the transfer energy, $F_{\max}$ is the contact force, $p_{\max}$ is the contact pressure, $k$ is the effective spring constant, $A$ is the contact area. $\mu$ is the reduced mass of the two-body system:
    \begin{equation}
        \mu = m_{r}^{-1}+m_{h}^{-1},
    \end{equation}
    where $m_{r}\in\mathbb{R}$ is the apparent mass of the robot, i.e. the mass perceived by the human operator during the collision and it can be computed as illustrated in \cite{pupaenergy}, and $m_{h}\in\mathbb{R}$ is the mass of the human body part, which is defined in the ISO/TS 15066.
    By combining the Speed and Separation Monitoring and Power and Force Limiting paradigms \cite{lucci2020RAL, ferraguti2020control}, it is possible to ensure high efficiency when sufficiently distant from the operator while permitting contact during the physical handover phase. This can be expressed as an upper bound on the maximum robot speed towards the human operator:
    \begin{equation}
        v_{rh}(t)\leq \max(v_{rh,SSM}(t),v_{rh,PFL}(t)),
        \label{eq:vmax}
    \end{equation}
    where $v_{rh}(t)$ represents the component of the robot velocity projected towards the human operator at time $t$.
    Since equation \eqref{eq:vmax} imposes an upper bound on the velocity, it is convenient to apply the path-velocity decomposition technique:
    \begin{align}
        & &\traj &= \trajs &&t\in \begin{bmatrix}t_i, t_f\end{bmatrix},
        \label{eq:q_s}
        \\
        & &\dtraj &= \dtrajs\dot{s} &&t\in \begin{bmatrix}t_i, t_f\end{bmatrix},    
        \label{eq:q_s_dot}
    \end{align}
    where the derivative $\dot{s}$ of the curvilinear abscissa $s$ parameterizes the geometric path $\trajs$ while preserving the overall path integrity.
    The trajectories thus obtained must be followed in a compliant manner with the operator, for example implementing admittance control, dynamically responding to external stimuli acting on the force sensor during the physical handover phase to result in a more natural and human-like movement.

    Lastly, once the trajectory has been computed, it is necessary to detect when the moment in which the human operator grabs the object and release it. The robot must ensure that the object is firmly in the grasp of the operator, while also being robust to potential disturbances or inadvertent impacts. These impacts may happen while the human operator tries to locate the object with the hand while not looking at the robot.
    % Lastly, once the trajectory has been computed, it is necessary to detect when the moment in which the human operator grabs the object and release it. The robot must ensure that the object is securely in the grasp of the operator, while also being robust to potential disturbances or inadvertent impacts. These impacts may happen while the human operator tries to locate the object with the hand while not looking at the robot.

    Therefore, the aim of this work is to develop a full architecture that:
    % To achieve this, we propose a compliant blind handover architecture that:
    \begin{itemize}
        \item Ensures safety throughout the handover procedure, while aiming to maximize efficiency.
        \item Recognizes the physical handover phase to automatically release the object when firmly held by the operator.
        % \item Recognizes the physical handover phase to automatically release the object when securely held by the operator.
        \item Demonstrates robustness to external impacts or disturbances, given the operator's lack of visual feedback during the blind handover.
    \end{itemize}

    \section{Compliant Handover Controller}\label{sec:admittance_controller}

    % \begin{itemize}
    %     \item present the proposed architecture and all the main components
    %     \item trajectory planning with quintic polynomial functions - cite corke toolbox
    %     \item admittance controller to follow the planned trajectory by ensuring compliance with the user during the contact during the handover phase
    % \end{itemize}

\begin{figure}[htbp]
        \centering
        \includegraphics[trim={8.5cm 2.5cm 8.5cm 1.5cm}, clip, width=\linewidth]{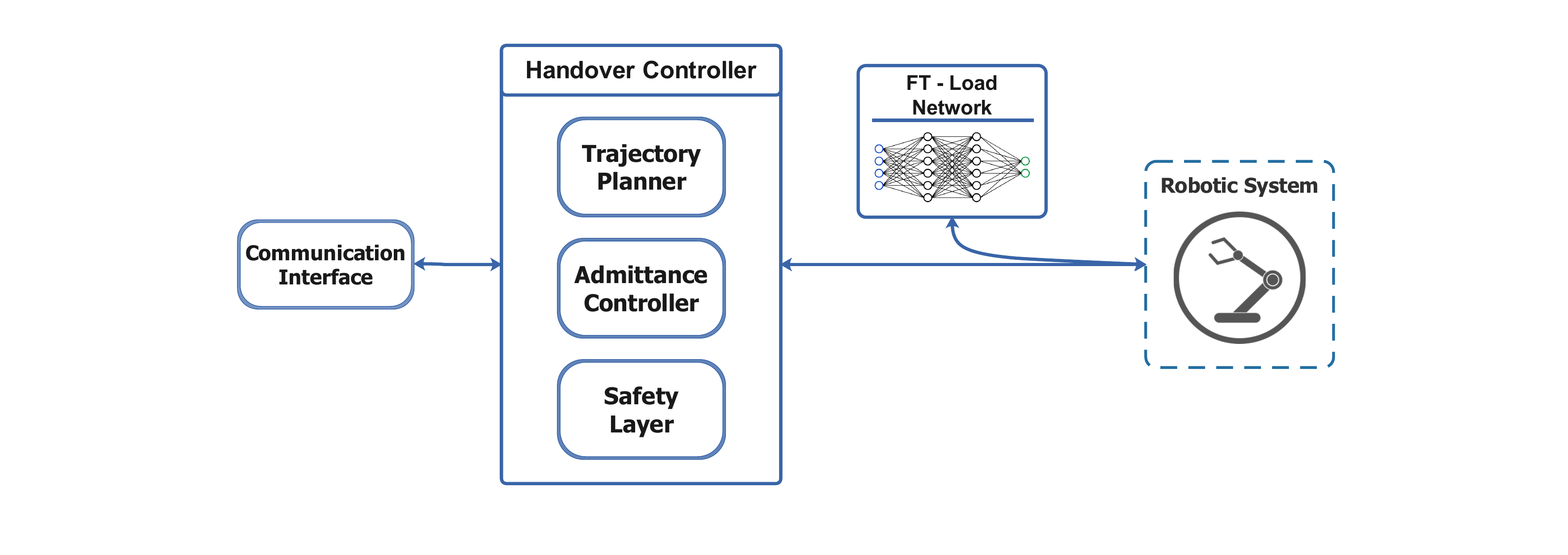}
        \caption{Proposed Architecture}
        \label{fig:architecture}
    \end{figure}

    The proposed architecture\footnote{GitHub: \href{https://github.com/davideferrari95/blind_handover_controller}{https://github.com/davideferrari95/blind\_handover\_controller}}, in Fig. \ref{fig:architecture}, comprises three main components: a human-robot communication interface, a controller for the handover procedure, and a neural network for detecting the object's load transfer during the physical handover.

    % \begin{figure*}
    % % [htbp]
    %     \centering
    %     \includegraphics[trim={4cm 2.5cm 8.5cm 1.5cm}, clip, width=0.8\linewidth]{Figures/Overall Architecture.pdf}
    %     \caption{Proposed Architecture}
    %     \label{fig:architecture}
    % \end{figure*}

    The communication interface manages operator requests and potential responses. In this architecture, we implemented a vocal communication channel, allowing the operator to naturally request the transfer of the desired object. This request is directed to the central handover controller that first plans the desired trajectory, designed to execute the pick of the requested object and deliver it to the operator. 
    % Once the trajectory is planned, an admittance controller calculates the desired velocity, by integrating over the controller rate $T_r$:
    Once the trajectory is planned, an admittance controller calculates the desired velocity $\dx_{adm}$ by integrating \eqref{eq:admittance}, and obtains the desired joint velocity $\dq_{adm}$ by pre-multiplying with the inverse of the robot's Jacobian matrix $J(\q)$.

    % \begin{equation}
    %     \ddot{x} = M^{-1} (D (\dot{x}_{des} - \dot{x}) + K (x_{des} - x) - F_{ext}) + \ddot{x}_{des}
    %     \label{eq:admittance}
    % \end{equation}

    \begin{equation}
        \ddx_{adm} = M^{-1} (D (\dx_{des} - \dx) + K (\x_{des} - \x) - \mathbf{F}_{ext}) + \ddx_{des}
        \label{eq:admittance}
    \end{equation}

    % \begin{equation}
    %     \dot{x} = \dot{x} + \ddot{x} \cdot T_r, \quad \dot{q} = J^{-1} \dot{x}
    %     \label{eq:admittance_integration}
    % \end{equation}

    % \begin{equation}
    %     \dot{x}_{k+1} = \dot{x}_k + \ddot{x}_k \cdot T_r, \quad \dot{q} = J^{-1} \dot{x}
    %     \label{eq:admittance_integration}
    % \end{equation}

    % \begin{equation}
    %     % \dot{x}_{k+1} = \dot{x}_k + \int_0^{T_r}{\ddot{x}_k \cdot dt}, \quad \dot{q} = J^{-1} \dot{x}
    %     \dot{x} = \dot{x} + \int_t^{t + T_r}{\ddot{x} \cdot dt}, \quad \dot{q} = J^{-1} \dot{x}
    %     \label{eq:admittance_integration}
    % \end{equation}

    \begin{equation}
        % \dot{x}_{k+1} = \dot{x}_k + \int_0^{T_r}{\ddot{x}_k \cdot dt}, \quad \dot{q} = J^{-1} \dot{x}
        % \dot{x} = \dot{x} + \int_t^{t + T_r}{\ddot{x} \cdot dt}, \quad \dot{q} = J^{-1} \dot{x}
        % \dot{q} = u = J^{-1}(q) \cdot \dot{x}
        \dq_{adm} = J^{-1}(\q) \cdot \dx_{adm}
        \label{eq:admittance_integration}
    \end{equation}
    where $M$, $D$, and $K$ are the mass, damping, and stiffness matrices, respectively, representing the implemented control dynamics for admittance. $\x_{des}$, $\dx_{des}$, and $\ddx_{des}$ denote the desired Cartesian position, velocity, and acceleration, obtained from the inverse kinematic of the planned trajectory.
    % obtained from trajectory planning.
    $\x$ and $\dx$ represent the current position and velocity of the robot, while $\mathbf{F}_{ext}$ signifies the vector of external forces read by the robot's force-torque sensor. 
    % $J$ is the robot's Jacobian, and $\dot{q}$ represents the joint velocity vector.
    The use of an admittance controller enables the robot to be compliant during the physical handover, ensuring a smoother and more natural interaction with the operator.
    Subsequently, a safety layer computes velocity limits according to the SSM and PFL paradigms of ISO/TS 15066 and formulates an optimization problem to determine a velocity scaling coefficient $\alpha$, ensuring compliance with the regulatory speed limits.
    Algorithm \ref{alg:controller} shows the controller steps. Please note that we used the Euler method for integration and, for ease of notation, a linear parametrization, i.e. $\ddot{s}=0$ .

    \begin{algorithm}[htbp]
        \small
        \begin{algorithmic}[1]
            \Require Requested Object $Obj$
            % \Ensure Handover
            \\
            \State $\mathcal{T} \gets$ Plan Trajectory
            \State $\mathcal{T}_s \gets \mathcal{T}$ Spline Interpolation
            \\
            \While{True}
                \\
                % \State $s = s + \alpha / T_r$ \hspace*{0pt}\hfill Increase Current Parameterization
                \State $s \gets integrate(s, \alpha, T_r)$ \hspace*{0pt}\hfill Increase Parameterization
                \State $\q_{des},\dq_{des},\ddq_{des} \gets \mathcal{T}_s(s)$ \hspace*{0pt}\hfill Get From Spline
                \State $\x_{des},\dx_{des},\ddx_{des} \gets \q_{des},\dq_{des},\ddq_{des}$ \hspace*{0pt}\hfill Kinematic
                \\
                % \State $\ddot{x}_{k+1} = M^{-1} (D (\dot{x}_{k, des} - \dot{x}_k) + K (x_{k, des} - x_k) - F_{ext}) \linebreak \hspace*{12mm} + \ddot{x}_{k, des}$
                % \State $\ddot{x}_{k+1} = M^{-1} (D (\dot{x}_{k+1, des} - \dot{x}_k) + K (x_{k+1, des} - x_k) \linebreak \hspace*{12.5mm} - F_{ext}) + \ddot{x}_{k, des}$
                % \\
                % \State $\dot{x}_{k+1} = \dot{x}_k + \ddot{x}_{k+1} \cdot T_r, \quad \dot{q}_{k+1} = J^{-1} \dot{x}_{k+1}$

                % \State $\dot{q} \gets \dot{x}$
                \State $\ddx_{adm} = M^{-1} (D (\dx_{des} - \dx) + K (\x_{des} - \x) - \mathbf{F}_{ext}) + \ddx_{des}$
                \State $\dx_{adm} \gets integrate(\ddx_{adm},\dx,T_r)$
                \State $\dq_{adm} = J^{-1}(\q) \dx_{adm}$
                % \State $\dot{x} = \dot{x} + \ddot{x} \cdot T_r, \quad \dot{q} = J^{-1} \dot{x}$
                \\
                \State $SSM_{lim}, PFL_{lim}$ $\gets$ $computeSafetyLimits()$
                \State $v_{lim} = \max(SSM_{lim}, PFL_{lim})$
                \\
                \State $\alpha \gets optimize(\dq_{adm}, v_{lim})$ \hspace*{0pt}\hfill  Optimal Scaling Factor
                % \State $\dot{x}_{k+1} = \alpha \cdot \dot{x}_{k+1}$ Scale Desired Velocity
                % \State $\dx \gets \alpha \cdot \dx$ \hspace*{0pt}\hfill  Velocity Scaling
                \State $\uin = \alpha \cdot \dq_{adm}$ \hspace*{0pt}\hfill  Velocity Scaling
                % \State $\dx = \alpha \cdot \dx$ Scale Desired Velocity
                \\
            \EndWhile
        \end{algorithmic}
        \caption{Handover Controller Algorithm}
        \label{alg:controller}
    \end{algorithm}

    Finally, an LSTM-based neural classifier continuously monitors readings from the robot's force sensor. This classifier recognizes patterns indicating the object's passage during the physical handover and signals the manipulator to release the object safely. This comprehensive architecture aims to provide a robust, efficient, and safe human-robot blind handover in HRC scenarios.

    \section{Safety Layer}\label{sec:safety_layer}

    % \begin{itemize}
    %     \item safety ISO/TS 15066 controller with SSM + PFL switch safety paradigms and scaling optimization problem - paper andrea
    % \end{itemize}

    The safety layer ensures that the robot behavior complies with safety standards. To achieve this, it modulates online the desired velocity of the admittance controller $\dq_{adm}$ \cite{pupa2021safety} solving online the following optimization problem:
    \begin{equation}
        \label{eq:problem}
        \begin{array}{ll@{}ll}
            \displaystyle \min_{\alpha} & -\alpha,                                                                                                                      \\[0.2cm]
            \text{s.t.}                                                                                                                                               \\[0.2cm]
            & \displaystyle J_{r_i}({\q})\dq_{adm}\alpha \leq v_{max_i}        \quad \forall i \in \{1,\dots,n\}, \\[0.2cm]
            & \displaystyle \dq_{min} \leq \dq_{adm}\alpha \leq \dq_{max},                                                                \\[0.2cm]
            & \displaystyle \ddot{\q}_{min}\leq \dfrac{\dq_{adm}\alpha - \dq}{T_r} \leq \ddot{\q}_{max},                                   \\[0.2cm]
            & \displaystyle 0 \leq \alpha \leq 1.                                                                                           \\[0.2cm]
        \end{array}
    \end{equation}
    Here, $\alpha \in \begin{bmatrix}0, 1 \end{bmatrix}$ is the optimization variable representing the scaling factor. $J_{r_i}(\q)\in \mathbb{R}^{1 \times n}$ is a modified Jacobian considering only the scalar velocity of the $i$-th link towards the human operator \cite{pupa2021safety}. $v_{max_i}$ is the velocity limit imposed by the combination of SSM and PFL safety paradigms, as expressed in equation \eqref{eq:vmax}. $\dq_{min} \in \mathbb{R}^{n}$ and $\dq_{max} \in \mathbb{R}^{n}$ are the joint velocity lower and upper bounds, respectively. $\ddot{\q}_{min} \in \mathbb{R}^{n}$ and $\ddot{\q}_{max} \in \mathbb{R}^{n}$ represent the acceleration limits. $\dq \in \mathbb{R}^{n}$ is the actual robot velocity, and $T_r$ is the robot execution time.
    The first constraint guarantees that the robot behaviour is always compliant with the safety standards, expressed as the maximum admissible velocity between the SSM and the PFL.
    The second and third constraints, represent the velocity and acceleration bounds of the system.

    The optimization problem in \eqref{eq:problem} is convex, i.e. it can be easily solved in real-time exploiting standard solvers. Furthermore, 
    % it has always a feasible solution greater than 0, i.e. \mbox{$\alpha = \min_{i \in \{1, \dots, n\}} \left( \frac{v_{\max,i}}{J_{r_{i}} \Delta q_{\text{adm}}} \right)$}.
    % Indeed, 
    differently from \cite{pupa2021safety}, by combining the both the SSM and the PFL it is possible to achieve a less conservative behaviour, i.e. the velocity of the robot is scaled less then required by a single collaborative paradigm. This allows to improve the overall performances.

    \section{Force-Load Transfer Network}\label{sec:force-load}

    % \begin{itemize}
    %     \item cite force-load curve transfer in human-human handover paper.
    %     \item describe and show FL-Transfer user study, with average FL-Curve
    %     \item describe the dataset creation, number, data...
    %     \item describe the NN implementation and training hyperparameters.
    % \end{itemize}

    % BATCH_SIZE, PATIENCE, LOAD_VELOCITIES, DISTURBANCES = 256, 50, False, True
    % HIDDEN_SIZE, SEQUENCE_LENGTH, STRIDE, OPEN_GRIPPER_LEN = [512, 256], 500, 10, 100
    % BALANCE_STRATEGY = ['weighted_loss', 'oversampling', 'undersampling']
    
    % \begin{figure*}
    %     \centering
    %     \includegraphics[width=\linewidth]{Figures/Questionnaire Results.png}
    %     \caption{Questionnaire Results}
    %     \label{fig:Questionnaire_Results}
    % \end{figure*}
    
    During the Physical Handover Phase, which involves the interaction between the giver and receiver during the object transfer, the possession and stability control of the object transition from the giver to the receiver. Recognizing the timing at which the robot should release the object is crucial to ensuring that the operator has stable control, preventing errors and inadvertent falls due to impacts or mistakes.
    % 
    % In this specific scenario, where we opted not to implement any visual feedback \cite{Yang21ICRA} or other wearable tactile sensors \cite{Leal19ISMCR}, reliance is placed on the robot's force/torque sensor to identify patterns indicating the handover. As highlighted in  \cite{Mason05EBR, Chan12HRI, Chan14ICRA}, feedback from a force sensor on the robot's wrist robustly modulates the release of an object. The handover process involves a gradual transition from the robot supporting the full load of the object to the receiver taking control. At the initiation of the handover, the robot supports the full load of the object $f_{L_0}$ by applying a grip force of $f_{G_0}$. As the Receiver starts taking the object, $f_L$ on the robot's hand decreases. The controller decreases the applied grip force $f_G$ accordingly in a linear fashion. When the Load Force reaches Zero, the controller maintains a small Non-Zero amount of Grip Force $f_{ZLG}$. Upon detecting a Slight Pull from the Receiver, the Load Force reaches a small Negative Value $\epsilon$, prompting the controller to release the object completely.
    %
    In this specific scenario, as we opted not to implement any visual feedback or other wearable tactile sensors, reliance is placed on the robot's force/torque sensor to identify patterns indicating the handover. As highlighted in \cite{Mason05EBR, Chan12HRI, Chan14ICRA}, feedback from a force sensor on the robot's wrist robustly modulates the release of an object.
    
    \begin{figure}[h]
        \centering
        \includegraphics[width=1\linewidth]{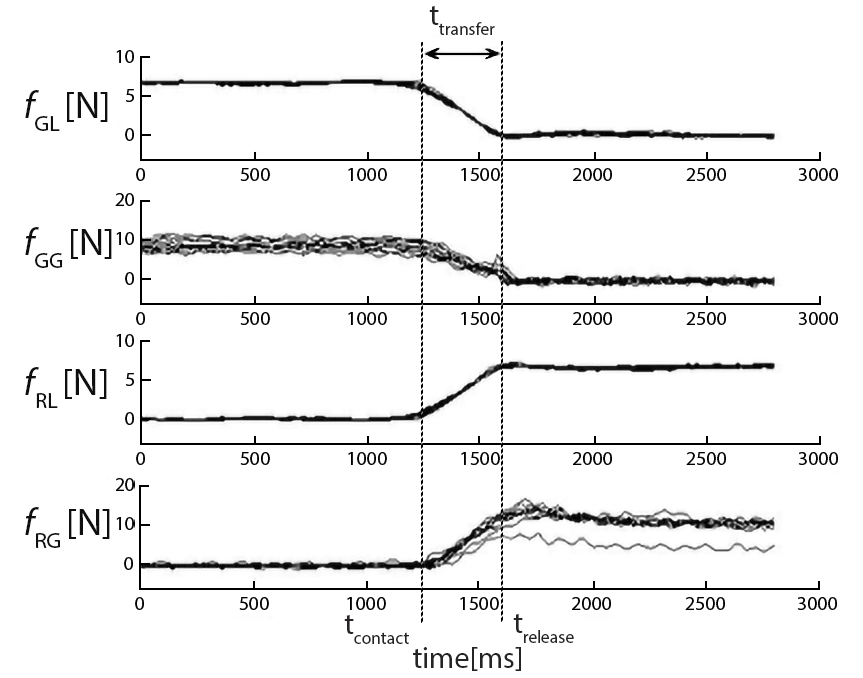}
        \vspace{-5mm}
        \caption{Load-Force Transfer Curve. $f_{GL}$, $f_{GG}$, $f_{RL}$, $f_{RG}$ represent load curve and grip force of the giver and receiver, respectively.}
        \label{fig:load-force-curve}
    \end{figure}

\begin{figure}[htbp]
        \centering
        \includegraphics[trim={1cm 7cm 20cm 5cm}, clip, width=\linewidth]{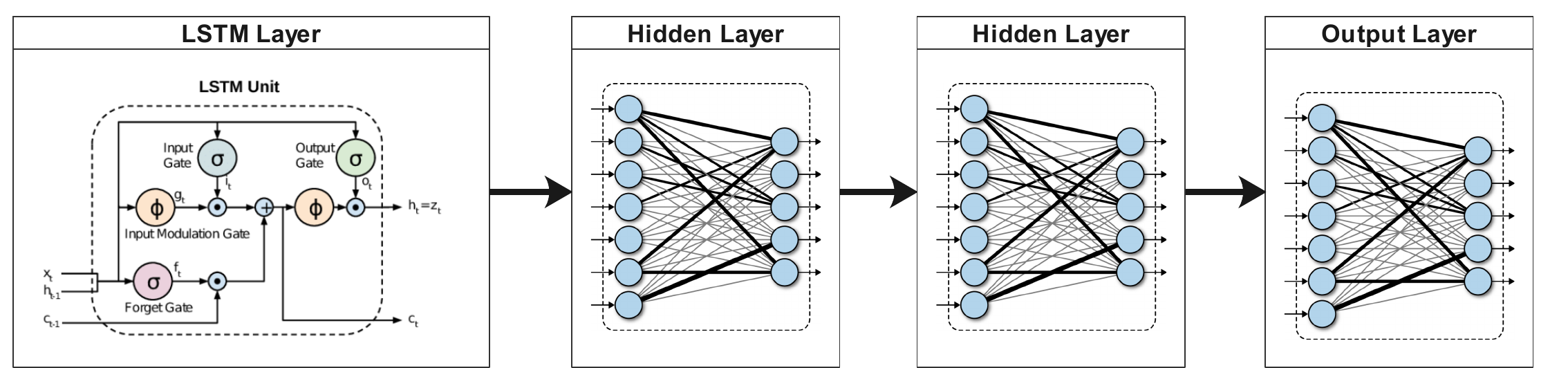}
        \caption{LSTM Neural Network}
        \label{fig:network}
    \end{figure}

    The handover process involves a gradual transition from the robot supporting the full load of the object to the receiver taking control, as shown in Fig. \ref{fig:load-force-curve}.
    The handover starts with the robot supporting the full load of the object $f_{L_0}$, by applying a grip force of $f_{G_0}$. As the Receiver starts taking the object, $f_L$ on the robot's hand decreases. The controller decreases the applied grip force $f_G$ accordingly in a linear fashion. When the Load Force reaches Zero, the controller maintains a small Non-Zero amount of Grip Force. Upon detecting a Slight Pull from the Receiver, the Load Force reaches a small Negative Value, prompting the controller to release the object completely.
    The proposed approach involves recognizing the force-load curve variation patterns of the robot, that represent the giver, using a neural network that monitors the robot's force sensor's latest 500 readings (approximately 1 second). The neural network in Fig. \ref{fig:network} comprises an LSTM layer \cite{hochreiter1997long} with an input size equal to the length of the input sequences (500), two hidden layers with 512 and 256 neurons, respectively, interspersed with Rectified Linear Unit (ReLU) \cite{agarap2018deep} activation functions, and an output layer of size 2 followed by a Sigmoid activation function.
     
    Training was performed with batches of 256 samples, Adam optimizer with a learning rate of 4e-3, utilizing an early stopping function with patience set to 50 epochs, monitoring the training binary cross-entropy loss \cite{Mao23ICML} trend:

    \begin{equation}
        \small
        L_{BCE} = - \frac{1}{n} \sum^n_{i=1}{(Y_i \cdot \log\hat{Y}_i + (1-Y_i) \cdot \log(1 - \hat{Y}_i))}
        \label{eq:BCE_loss}
    \end{equation}

    \begin{figure*}[b]
        \centering
        \includegraphics[width=\linewidth]{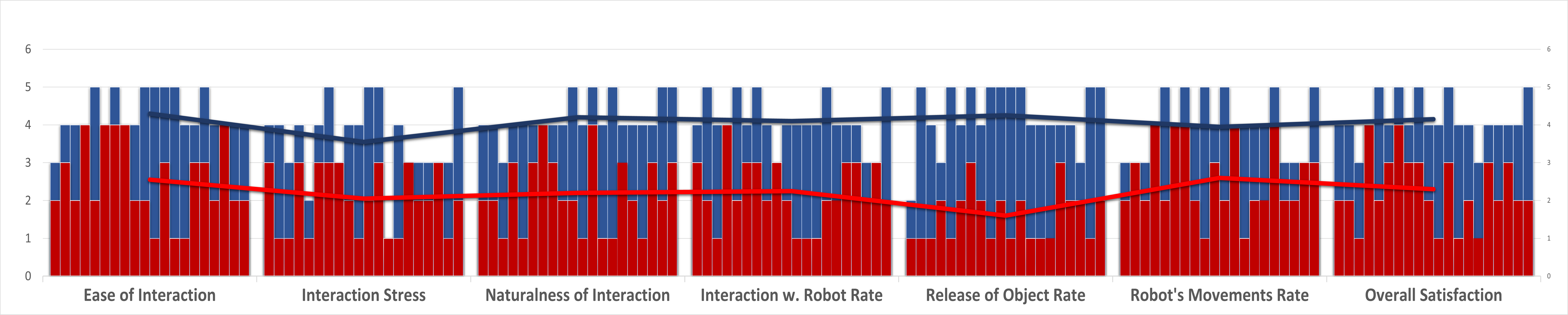}
        \caption{Questionnaire Results}
        \label{fig:Questionnaire_Results}
    \end{figure*}
    
    To create the dataset to train the network, we conducted a user study, collecting 300 handover examples with various objects, grasping poses, handover poses, and human movements. For each example, a sequence of force sensor readings at 500Hz was saved, paired with a label for each row (1 if the gripper should open, 0 otherwise). These sequences were pre-processed to create the neural network's training dataset, generating samples by sliding a window of 500 consecutive readings with a stride of 10 elements. Each sample was associated with the label corresponding to the last row (i.e., if you recognize this sequence, you should perform the action of the last instant of the sequence), indicating whether the gripper should open or not.
    This resulted in creating more than 200,000 samples for the neural network. Dataset balancing techniques such as \textit{weighted loss}, \textit{oversampling}, and \textit{undersampling} were applied to achieve a homogeneous dataset for both '\textit{open}' and '\textit{do not open}' gripper classes. Furthermore, we collected additional samples where disturbances were simulated in the form of inadvertent impacts on the object or the force sensor itself. This trained the network to be robust against potential undesired interactions, as the user, being faced in another direction, might fumble to locate the object and choose the optimal grasping pose blindly.
    During runtime, each new force sensor reading is added to a FIFO buffer representing the sequence of the last n readings. The buffer is encoded into a tensor and passed through the neural network to obtain the output indicating whether to open the gripper or not, in order to complete the physical handover phase.

\section{Experimental Validation}\label{sec:exp}
    
    % \begin{itemize}
    %     \item experimental validation description
    %     \item user study, evaluation metrics... 
    % \end{itemize}

    % \begin{figure*}
    %     \centering
    %     \includegraphics[width=\linewidth]{Figures/Questionnaire Results.png}
    %     \caption{Questionnaire Results}
    %     \label{fig:Questionnaire_Results}
    % \end{figure*}
    
    % The experimental validation of our blind handover architecture included a comparative study with a state-of-the-art approach, where the robot performs the handover procedure using a PD controller without compliance control with the operator and with a force threshold to detect the gripper's opening moment.
    The experimental validation of our blind handover architecture included a comparative study with a state-of-the-art approach, where:
    \begin{enumerate}
        \item The robot performs the handover procedure using a PD controller without compliance control with the operator.
        \item Uses a Force-Threshold to detect the gripper's opening moment, as described in \cite{Chan14ICRA}.
    \end{enumerate}
    The study encompassed 20 participants aged between 20 and 30, exhibiting various levels of experience with robotic systems, ranging from novices to individuals with several years of experience with collaborative robots. This diverse participant pool aimed to ensure a representative sample. To mitigate potential learning biases, each participant engaged in both versions of the experiment, and the execution order was randomly determined for each individual, minimizing the impact of experiment repetition on the results.
    % 
    % \begin{figure}[htbp]
    \begin{figure}[htbp]
        \centering
        \vspace{2mm}
        \includegraphics[width=0.7\linewidth]{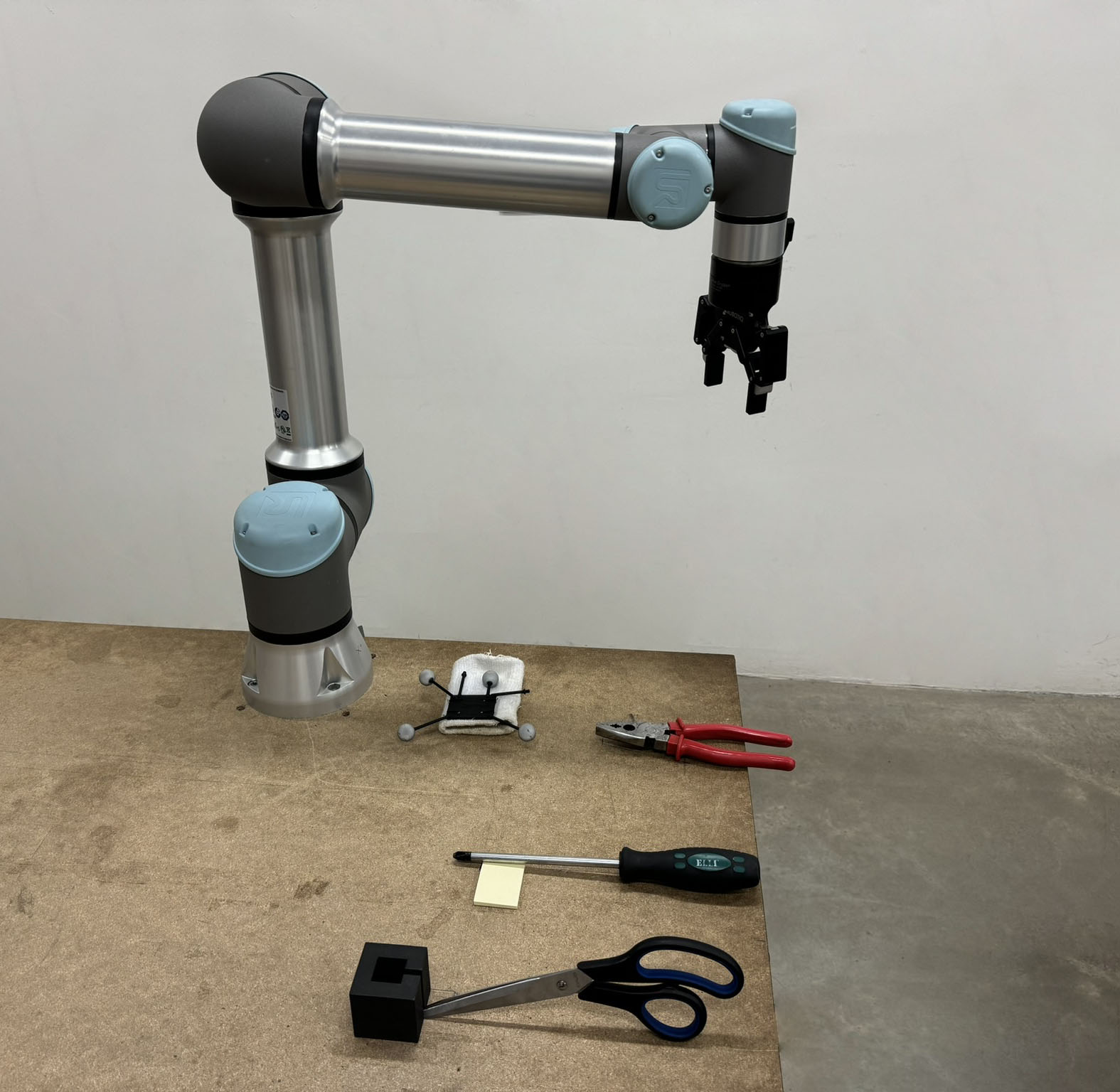}
        \caption{Setup of the Experiment}
        \label{fig:setup}
    \end{figure}
    In this experimental setup (Fig. \ref{fig:setup}), participants undertook a blind handover task, requesting and receiving three different tools from a UR10e collaborative manipulator. The robot, through a vocal communication interface, informed the operator that it was delivering the requested object and tracked the operator's hand to facilitate the handover. The operator blindly assumed control of the object, and the robot determined when to release the grip using the two different architectures.
    
    The primary objective of these experiments was to gather user experiences after interacting with the two different architectures and evaluate whether the use of a compliant movement with the operator during the physical handover, coupled with a neural network facilitating a natural and robust object transfer, could influence user satisfaction.
    
    The subsequent sections delve into a comprehensive examination of the implemented architecture and the experimental outcomes, offering insights into the advantages and improvements realized through our blind handover architecture.

    % \begin{figure*}
    %     \centering
    %     \includegraphics[width=\linewidth]{Figures/Questionnaire Results.png}
    %     \caption{Questionnaire Results}
    %     \label{fig:Questionnaire_Results}
    % \end{figure*}

    \subsection{Architecture Implementation}\label{subsec:implementation}
        
        % \begin{itemize}
        %     \item details on the communication (alexa) - paper irim
        %     \item details on ros and the architecture integration ?
        % \end{itemize}

        The architecture was developed within the \textit{ROS} framework \cite{ros}, with components organized into independent nodes to ensure modularity. For the voice communication channel, we adopted the architecture proposed in \cite{Ferrari23IRIM}, designed using \textit{Alexa Conversations} \cite{acharya2021alexa}, a Deep Learning-based approach that utilizes API calls to manage multi-turn dialogues between Alexa and the user, resulting in more natural and human-like interactions. To handle trajectory planning, we employ quintic polynomials using the Peter Corke Robotics Toolbox \cite{Corke21ICRA}, interpolating them with cubic splines to streamline the path velocity decomposition process and admittance calculation.
    
        % Given that compliance with ISO safety standards is ensured in both experiments, we aim to use these metrics to valuate how well our architecture can predict and prevent slowdowns or safety stops through communication, specifically those arising from excessive proximity to the operator.
        
    \subsection{Analysis of the Results}\label{subsec:results}

        To validate the proposed architecture, participants were presented with a user experience questionnaire, where they had to rate on a scale from 1 to 5 \textit{"ease of interaction with the robot"}, \textit{"stress during interaction"}, \textit{"naturalness of interaction"}, \textit{"interaction with the robot"}, \textit{"object release"}, \textit{"robot movement during handover"} and \textit{"overall satisfaction"} for both versions of the comparative experiment. The results of the questionnaire are depicted in Fig. \ref{fig:Questionnaire_Results}, with values from the standard experiment shown in red and those from the architecture proposed in this article shown in blue.         
        
        From the averages resulting from the analysis of the results, we can observe a significant difference in user preferences between the two architectures, especially regarding the \textit{naturalness of interaction} and the evaluation of \textit{object release}. This highlights the importance of ensuring the simplest and most natural possible physical handover, emphasizing how a neural network can be an excellent tool for achieving this. In general, we can highlight an average improvement of about 2 points out of 5 in favor of the proposed architecture, underscoring that the increased naturalness of the handover also affects the accumulated stress and overall process evaluations. Through an analysis of variance (ANOVA) (Tab. \textbf{\ref{tab:ANOVA_questionnaire}}), we verified the significance of the proposed results: obtaining an $F-value = 134.521 >> F_{crit} = 4.7472$, significantly exceeding the critical value, demonstrating the significance of the differences highlighted by the collected data.

        \begin{table}
            \vspace{5pt}
            \small
            \caption{ANOVA Summary and Results - Questionnaire}
            \label{tab:ANOVA_questionnaire}
            \centering
            \resizebox{0.8\linewidth}{!}{
            \begin{tabular}{c|c|c|c|c}
                \toprule
                \textbf{Groups} & \textbf{Count} & \textbf{Sum} & \textbf{Average} & \textbf{Variance} \\
                \midrule
                Standard & 7 & 15.55 & 2.22 & 0.11 \\
                Proposed & 7 & 28.5 & 4.07 & 0.06 \\
                \bottomrule
                \toprule
                \textbf{F} & \multicolumn{2}{c|}{\textbf{P-value}} & \multicolumn{2}{c}{\textbf{F crit.}} \\
                \midrule
                134.521 & \multicolumn{2}{c|}{7.06e-8} & \multicolumn{2}{c}{4.7472} \\
                \bottomrule
            \end{tabular}}
        \end{table}

        \begin{figure}[htbp]
            \centering
            \includegraphics[width=0.9\linewidth]{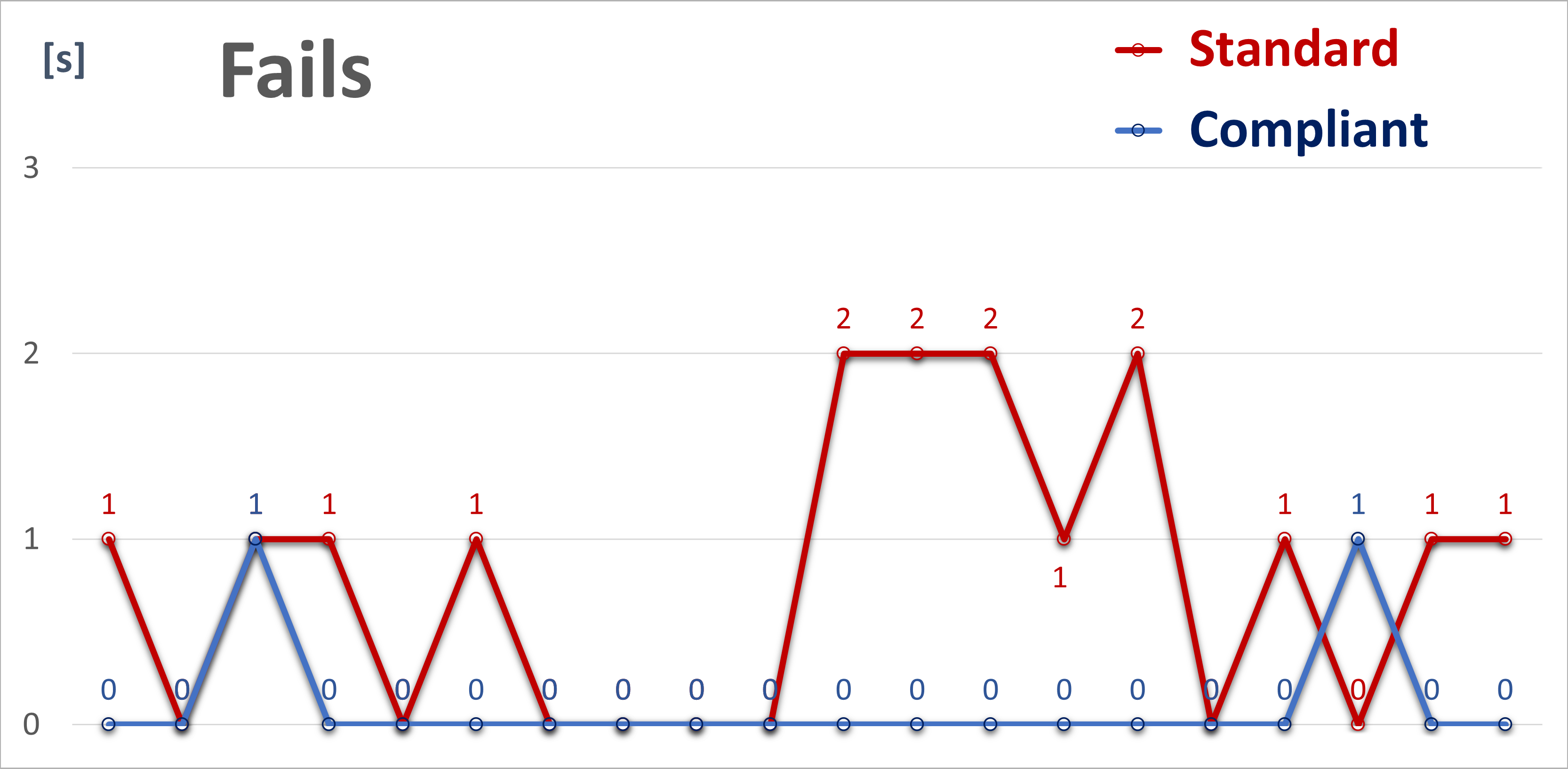}
            \caption{Handover Fails}
            \label{fig:fails}
        \end{figure}

        In addition, Fig. \ref{fig:fails} illustrates the number of errors encountered for the two architectures during the experiments. An error may represent the premature dropping of the object before or during the physical handover phase due to unintentional impacts or disturbances, or a failed or difficult release of the object, necessitating the interruption of the task execution. As depicted in the graph, the proposed architecture allows for minimizing the number of failures (only 2 errors out of 60) compared to the standard one (16 errors out of 60). This demonstrates how training the network to be robust to unintentional impacts has ensured a significant improvement in the final experiment performance.

    \section{Conclusions and Future Work}\label{sec:conclusions}

    % \begin{itemize}
    %     % \item train NN to understand on-the-fly handover. - DONE
    %     \item context aware grasp planning and handover pose for receiver task.
    %     \item error handling ? grip force modulation ? gripper tactile sensors ?
    % \end{itemize}

    In this paper, we introduced a innovative blind handover architecture designed to prioritize safety while enabling a robust and natural handover. Our architecture allows an operator engaged in a task requiring constant attention to request an object from an assistant robot without interrupting ongoing operations. The operator relies entirely on the robot's ability to bring the object to the extended hand and ensure a safe transfer, all while adhering to ISO safety regulations.
    The experimental validation, conducted through a user preference questionnaire, highlights the effectiveness of the architecture, resulting in a significant improvement in user experience. Additionally, it ensures a reduced occurrence of errors due to the specific reinforcement of the neural network to prevent such incidents.

    Future work will focus on implementing advanced computer vision algorithms to recognize the requested object in challenging situations and track the operator's hand using a true-depth camera mounted on the robot's end-effector. This would also enable the development of AI visual feedback to complement the neural network, further fortifying the physical handover phase and managing potential errors. This can also allow to conduct a study on optimal grasping and handover poses concerning both the object's position and the operator's future use, to make the blind handover more efficient and natural, also considering crucial manipulability problems, such as slippage, to optimize the grasping strategy based on the shape of the gripper and the desired object.
    Moreover, there is potential to implement a more advanced communicative structure, as seen in \cite{Ferrari2023HFR}, by incorporating the simultaneous management of multiple communication channels to expand interaction possibilities between the operator and the robot.

    % Bibliography
    \bibliographystyle{IEEEtran}
    \bibliography{mybib.bib}
    
\end{document}